\documentclass[10pt,twocolumn,letterpaper]{article}

\usepackage{iccv}              % To produce the CAMERA-READY version
\usepackage[accsupp]{axessibility} % To Improves PDF readability
\newcommand{\myparagraph}[1]{\vspace{1pt} \noindent \textbf{#1} \ }
\definecolor{iccvblue}{rgb}{0.21,0.49,0.74}
\usepackage[pagebackref,breaklinks,colorlinks,allcolors=iccvblue]{hyperref}
\usepackage{url}
\usepackage{graphicx}
\usepackage{subcaption}
\usepackage{multirow}
\usepackage{arydshln}
\usepackage{amsmath} 
\usepackage[normalem]{ulem}

\usepackage{xcolor}
\usepackage{tcolorbox}
\tcbuselibrary{listings, breakable}
\newtcblisting{prompt}{
  breakable,
  colback=gray!5,
  colframe=gray!75!black,
  fonttitle=\bfseries,
  title={Prompt for Detailed Subject Caption},
  listing only,
  listing options={
    basicstyle=\small\ttfamily,
    breaklines=true,
    keepspaces=true,
    showstringspaces=false
  }
}

\useunder{\uline}{\ul}{}
\def\paperID{10984} % *** Enter the Paper ID here

\newcommand{\mengping}[1]{\textcolor{blue}{[Mengping: #1]}}

\def\confName{ICCV}
\def\confYear{2025}

\title{FreeCus: Free Lunch Subject-driven Customization in Diffusion Transformers}

\begin{document}

\author{
%
% Yanbing Zhang$^{1}$, Zhe Wang$^{1*}$, Qin Zhou$^{1*}$, Mengping Yang$^{2}$ \\
Yanbing Zhang$^{1,2}$, Zhe Wang$^{1,2*}$, Qin Zhou$^{1,2*}$, Mengping Yang$^{3}$ \\
% Yanbing Zhang$^{1,2}$, Zhe Wang$^{1,2*}$, Qin Zhou$^{1,2*}$, Mengping Yang$^{1,2}$ \\
\textsuperscript{1} Key Laboratory of Smart Manufacturing in Energy Chemical Process, ECUST, China\\
\textsuperscript{2} Department of Computer Science and Engineering, ECUST, China\\
% \textsuperscript{2} Key Laboratory of Smart Manufacturing in Energy Chemical Process, ECUST, China\\
\textsuperscript{3} Shanghai Academy of Al for Science, China\\
{\tt\small zhangyanbing@mail.ecust.edu.cn, \{wangzhe, sunniezq\}@ecust.edu.cn, kobeshegu@gmail.com}
}

\twocolumn[{%
    \renewcommand\twocolumn[1][]{#1}%
    \maketitle
    \begin{center}
        \centering
        \vspace{-25pt}
        \includegraphics[width=\linewidth]{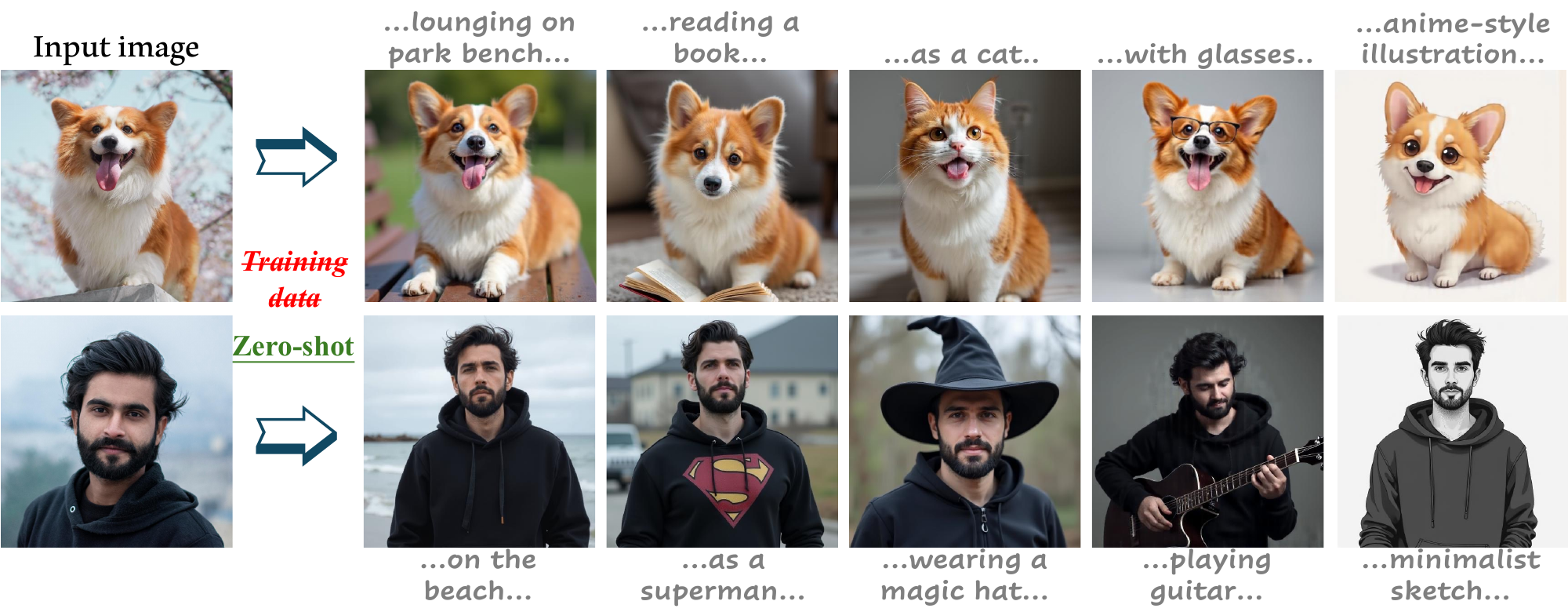}
        \vspace{-10pt}
        \setlength{\abovecaptionskip}{-2pt}
        \captionof{figure}{ 
            % Referencing the given image from users, our proposed method can synthesize the consistent subject  while following flexible target prompts, without any training samples, optimized embedding or encoders.
            Given the user-provided image as a reference, our proposed method synthesizes a consistent subject while adhering to flexible target prompts, all without the need for training samples, optimized embeddings, or encoders.
        }
        \label{fig:teaser}
    \end{center}
    
}]

%   \includegraphics[width=\textwidth]{figures/teaser.pdf}

\maketitle
{
  \renewcommand{\thefootnote}%
    {\fnsymbol{footnote}}
  \footnotetext[1]{Corresponding author}
}
\begin{abstract}
% In light of the rapid advancements in text-to-image generation, especially recent diffusion transformers, a growing number of users are leveraging subject-driven technologies to synthesize high-quality images that closely resemble input images, thereby enhancing design processes and providing entertainment.
% In light of the rapid advancements in text-to-image generation, especially with recent diffusion transformers, an increasing number of users are employing subject-driven technologies to synthesize high-quality images that closely resemble input images, thereby enhancing design workflows and offering engaging entertainment.
%
% \mengping{
In light of recent breakthroughs in text-to-image (T2I) generation, particularly with diffusion transformers (DiT), subject-driven technologies are increasingly being employed for high-fidelity customized production that preserves subject identity from reference inputs, enabling thrilling design workflows and engaging entertainment.
% identify
%
Existing alternatives typically require either per-subject optimization via trainable text embeddings or training specialized encoders for subject feature extraction on large-scale datasets.
Such dependencies on training procedures fundamentally constrain their practical applications.
More importantly, current methodologies fail to fully leverage the inherent zero-shot potential of modern diffusion transformers (\emph{e.g.,} the Flux series) for authentic subject-driven synthesis.
To bridge this gap, we propose \textbf{FreeCus}, a genuinely training-free framework that activates DiT's capabilities through three key innovations:
1) We introduce a pivotal attention sharing mechanism that captures the subject's layout integrity while preserving crucial editing flexibility.
% systematic/straightforward
2) Through a straightforward analysis of DiT's dynamic shifting, we propose an upgraded variant that significantly improves fine-grained feature extraction.
3) We further integrate advanced Multimodal Large Language Models (MLLMs) to enrich cross-modal semantic representations.
Extensive experiments reflect that our method successfully unlocks DiT's zero-shot ability for consistent subject synthesis across diverse contexts, achieving state-of-the-art or comparable results compared to approaches that require additional training.
Notably, our framework demonstrates seamless compatibility with existing inpainting pipelines and control modules, facilitating more compelling experiences.
Our code is available at: https://github.com/Monalissaa/FreeCus.
\end{abstract}

\section{Introduction}
Nowadays, text-to-image (T2I) models \cite{sd3, flux} can generate photorealistic images that sometimes surpass the quality of real photographs. Leveraging these capabilities, users increasingly employ T2I models for image-to-image tasks \cite{instantstyle, nullText, styleAligned, deutch2024turboedit} in design and entertainment. Among these applications, subject-driven generation \cite{textualInversion, ruiz2023dreambooth, mao2024realcustom++}, also termed customization or personalization, has gained prominence for enabling contextually diverse image generation while maintaining subject consistency, as illustrated in Fig. \ref{fig:teaser}. This work focuses on \textit{training-free} subject-driven T2I generation, which circumvents additional training and remains underexplored due to challenges in aligning visual-text feature space without explicit training.  

Existing subject-driven methods fall into two groups. The first \cite{textualInversion, customDiffusion, huang2024incontext, zhang2024attention} fine-tunes base models using limited subject-specific samples (1–100), capturing unique features at the cost of laborious per-subject retraining. The second group \cite{msDiffusion, mao2024realcustom++, song2024moma} trains encoders on large-scale datasets (more than 10,000 samples with multi-view images) to align visual-text features, enabling training-free generalization across subjects. While avoiding retraining, this approach requires substantial computation and extensive sample collection for encoder training. Critically, neither paradigm achieves genuine zero-shot personalization.  

\begin{figure}[!t]
  \centering
  \includegraphics[width=\linewidth]{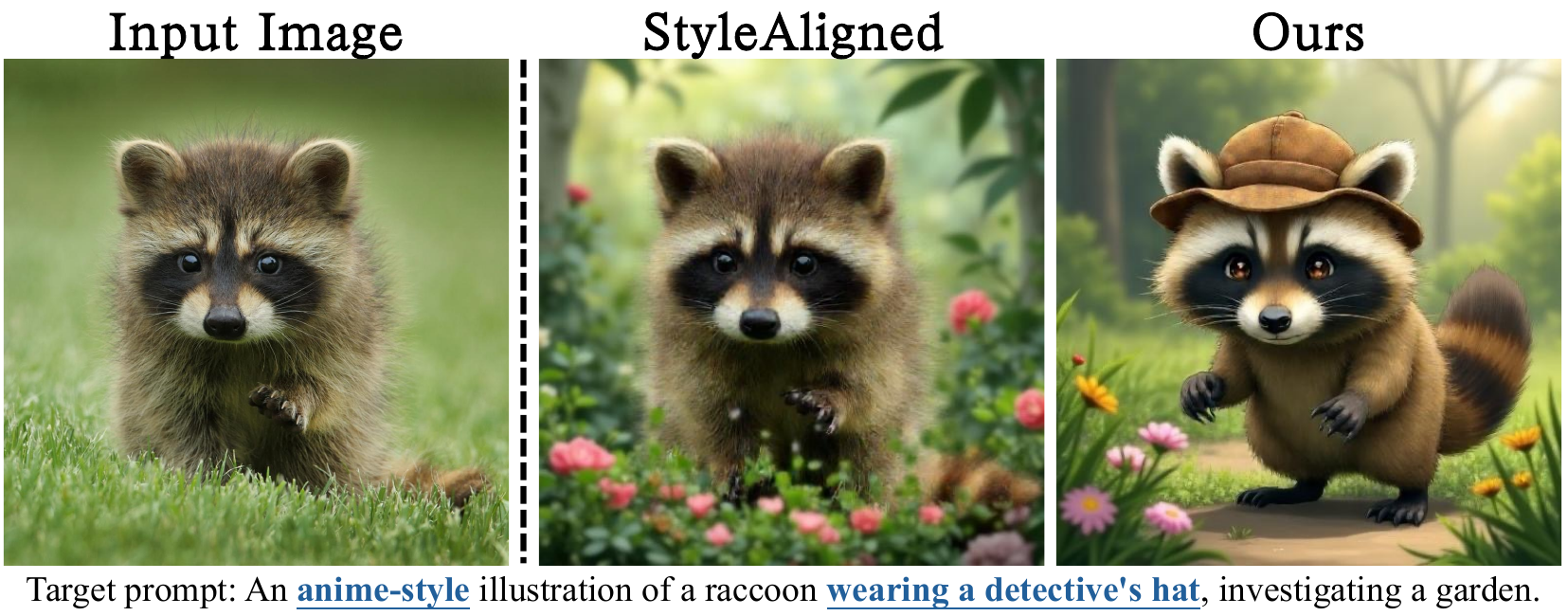}
  \caption{
  \textbf{Issues with StyleAligned \cite{styleAligned} in personalization.}
 Attention sharing causes text misalignment, e.g., it neither renders the anime style nor synthesizes the intended hat. Note: a mask is applied with StyleAligned to avoid fully replicating the input.
  }
  \label{fig:problems}
  \vspace{-15pt}
\end{figure}

Effectively integrating visual features from specific subjects into generated images is a key challenge for training-oriented subject-driven generation methods and simultaneously serves as the cornerstone for realizing our proposed training-free framework. 
Pretrained foundation models exhibit strong feature injection capabilities in zero-shot style transfer \cite{styleAligned}, inpainting \cite{avrahami2022blended, addit}, editing \cite{stableFlow}, and other layout-preserving tasks. Modern diffusion transformers (DiTs) \cite{sd3, flux} further outperform U-Net–based diffusion models. However, directly applying ``attention sharing'' mechanisms commonly used in layout preservation tasks significantly reduces editability (Fig. \ref{fig:problems}), failing to meet the flexibility demands of subject-driven generation (e.g., synthesizing anime styles or accessories in Fig. \ref{fig:teaser}).

 To address the decline in editability while preserving subject consistency, we propose a novel framework leveraging DiT’s zero-shot potential, named \textit{FreeCus}. First, we restrict attention sharing to critical DiT layers, which encode essential content features \cite{stableFlow}, enhancing text alignment and layout retention. Background regions, segmented via \cite{zheng2024birefnet}, are masked to minimize contextual interference. 
 Furthermore, the streamlined nature of attention sharing risks detail loss; thus, we adjust DiT’s dynamic shifting mechanism while extracting attention for the given subject. Finally, to compensate for incomplete semantic feature integration (e.g., color), we augment the framework with MLLM-derived information \cite{Qwen2VL, qwen2.5}.
 
  % Our contributions are summarized as follows: 
  % 1) A genuinely training-free subject-driven framework rivaling retraining-based methods; 2) Novel strategies including pivotal attention sharing, adjusted temporal shifting, and semantic enhancement to balance fidelity and controllability by fully exploiting DiT's zero-shot potential; 3) Compatibility with existing DiT-based methods and extensibility to tasks like inpainting and style transfer.

  % \mengping{
To sum up, our main contributions are: 
% your framework
1) We propose \textit{FreeCus}, a novel training-free framework for zero-shot subject-driven synthesis, fully leveraging pretrained DiT’s capabilities to generate consistent subjects in creative contexts;
% your components
2) An enhanced pivotal attention sharing mechanism, together with upgraded dynamic shifting and strategic integration of MLLMs, synergistically optimizing the balance between fidelity and controllability;
% flexibility and compatibility
3) The key ingredients of our framework are orthogonal and compatible with existing DiT-based models, and its versatile design enables seamless integration into other applications like style transfer and inpainting;
% Versatile design that extends to style transfer applications while remaining orthogonal and compatible with existing DiT-based methods; 
% experiment results and appealling applications
4) FreeCus achieves state-of-the-art performance in extensive comparisons, rivaling methods that require additional training.

\begin{figure*}[!t]
  \centering
  \includegraphics[width=\linewidth]{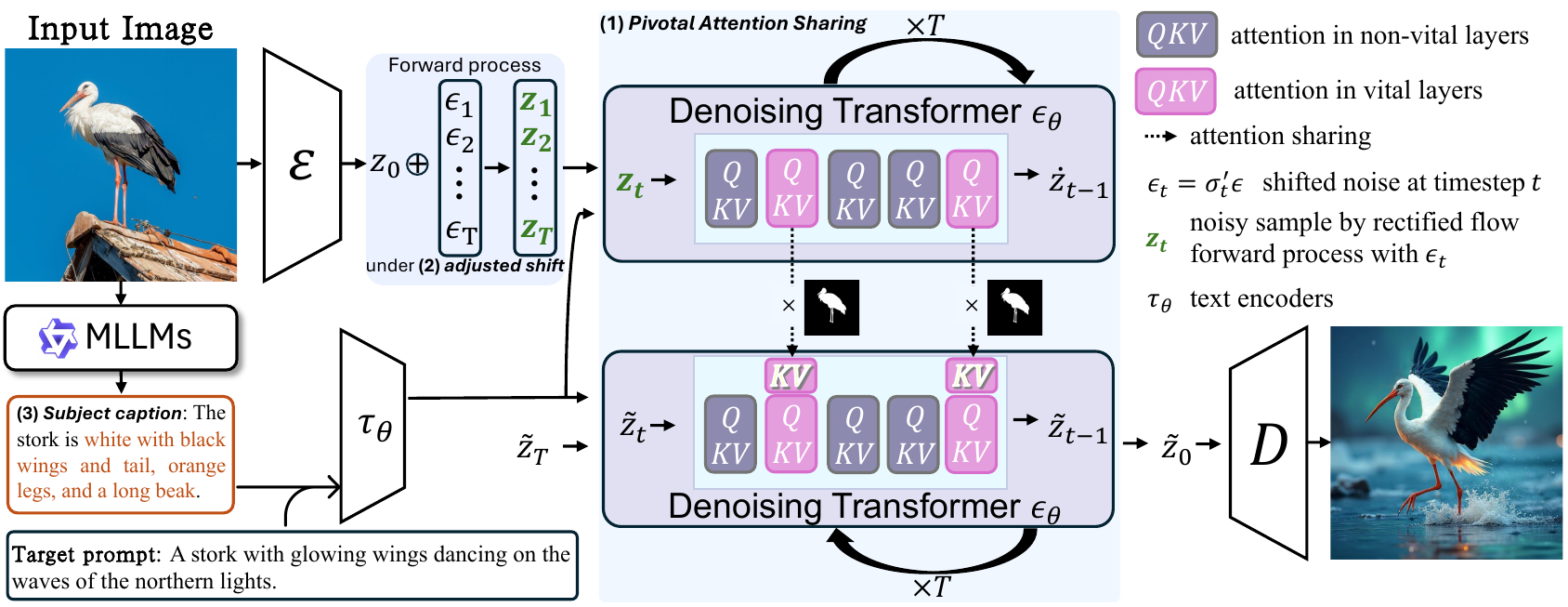}
  \caption{\textbf{Method overview}. Our approach transfers characteristics from a reference image $z_0$ to a target image $\tilde{z}_0$ through three mechanisms: (1) pivotal attention sharing, masking attention in critical layers to inject structural features while preserving editing flexibility; (2) adjusted dynamic shifting, deriving an improved diffusion trajectory $(z_1, ..., z_T)$ processed via rectified flow to enhance detail alignment between reference and target images; and (3) Multimodal LLM integration, extracting supplementary subject captions to capture semantic attributes potentially missed during attention sharing, thereby ensuring comprehensive subject representation.}
  % \caption{\textbf{Method overview}. Given a reference image $z_0$, we first derive attention by denoising the diffusion trajectory $z_1, ..., z_T$, which is processed through a rectified flow forward. This process utilizes a modified time schedule to enhance detail features in the attention. Next, we share the masked attention from critical layers during the generation of the target image $\tilde{z}_0$, effectively capturing the subject layout while preserving strong editing capabilities. Finally, we leverage multimodal LLMs to extract additional subject captions, supplementing semantic features of the subject that may have been overlooked during the attention sharing.} 
  \label{fig:overall}
  \vspace{-15pt}
\end{figure*}
\section{Related Work}

\subsection{Diffusion-based Text-to-Image Models}
% Diffusion-base T2I models \cite{diffusionBeatsGan, ho2020DDPM} have dominated the realm of image synthesis for past four years. In the early stage, they main denoise on pixel space \cite{balaji2022ediff,ho2022cascaded, nichol2021glide,saharia2022imagen}. For the diffusion model trained on the latent space, LDM \cite{rombach2022LDM} finds a good trade-off between the computational resources and generation quality. Then, a series of techniques are applied to improve performance, such as changing the text encoder \cite{radford2021clip, openclip}, improved autoencoder \cite{podellsdxl}, and cascaded architecture \cite{StableCascade}. Noticeable, DiT \cite{peebles2023DiT} explores diffusion models based on transformer \cite{vaswani2017transformer} architecture, instead of convolutional U-Net, and proving that diffusion transformers have the scaling potentiality as same as the LLMs \cite{brown2020gpt3, ouyang2022chatgpt}. After that, efficient training strategies \cite{chen2023pixartAlpha, chen2024pixartSigma}, flow-matching framework \cite{gao2024luminaT2x, zhuoluminaNext, liu2023flow}, multi-modal attention mechanism \cite{sd3, flux} are proposed to enhance the stability and scalability of DiT. Among these, Flux.1 \cite{flux} improves the rectified flow transformer with multi-modal attention to get extremely high generation quality and we exploit it to achieve the zero-shot subject-drive image generation.
Diffusion-based text-to-image (T2I) models \cite{diffusionBeatsGan, ho2020DDPM} have dominated the field of image synthesis for the past four years. In the early stages, these models primarily focused on denoising in pixel space \cite{balaji2022ediff, ho2022cascaded, nichol2021glide, saharia2022imagen}. The introduction of latent diffusion models (LDM) \cite{rombach2022LDM} established a favorable trade-off between computational resources and generation quality. Subsequently, various techniques were developed to enhance performance, including modifications to text encoders \cite{radford2021clip, openclip}, improvements in autoencoders \cite{podellsdxl}, and the adoption of cascaded architectures \cite{StableCascade}.
%
%Notably, the Diffusion Transformer (DiT) \cite{peebles2023DiT} explores diffusion models based on transformer architecture \cite{vaswani2017transformer}, moving away from traditional convolutional U-Nets, and demonstrating that diffusion transformers possess good scalability properties. 
Notably, the Diffusion Transformer (DiT) \cite{peebles2023DiT} investigates transformer-based diffusion models \cite{vaswani2017transformer}, replacing traditional convolutional U-Nets while demonstrating strong scalability.
Following this, efficient training strategies \cite{chen2023pixartAlpha, chen2024pixartSigma}, flow-matching frameworks \cite{gao2024luminaT2x, zhuoluminaNext, liu2023flow}, and multi-modal attention mechanisms \cite{sd3, flux} have been proposed to enhance the stability and scalability of DiT. Among these, Flux.1 \cite{flux} enhances the rectified flow transformer with multi-modal attention, improving generation quality. We leverage this framework to achieve zero-shot subject-driven image generation.
 
\subsection{Subject-Driven Image Generation}
Subject-driven image generation \cite{textualInversion, ruiz2023dreambooth, ye2023ipa, avrahami2023break, msDiffusion, chen2024anydoor} aims to synthesize images featuring a consistent subject in diverse contexts. Existing methods can be broadly divided into two categories based on whether they require retraining for every new subject.
For per-subject retraining, referred to as optimization-based customization, Textual Inversion \cite{textualInversion} optimizes a trainable token embedding using 3–5 user-provided images of the same object. Other methods \cite{ruiz2023dreambooth, customDiffusion, tewel2023key_locked} involve additional trainable parameters, particularly in the cross-attention layers. While these approaches demand relatively low computational resources, they are limited to fitting one subject at a time.
In contrast, optimization-free customization schemes leverage large-scale datasets to enable robust personalization without retraining for each new subject. Following the ideas introduced by textual inversion, several works \cite{li2024blipDiffusion, gal2023encoder, wei2023elite} also employ extra text embeddings while further training an auxiliary image encoder to map image features and update cross-attention weights.
%
% Distinct from these approaches, IP-Adapter \cite{ye2023ipa} argues that merging image and text features within the cross-attention layers can hinder fine-grained control, and thus introduces a lightweight adapter to decouple them. 
Distinct from these approaches, IP-Adapter \cite{ye2023ipa} argues that merging image and text features in cross-attention layers can hinder fine-grained control and proposes a lightweight adapter to decouple these features.
Other works \cite{patel2024lambda, pankosmos, song2024moma} employ multi-modal training to align image and text features better, and some further extract comprehensive subject features using multiple image encoders \cite{kong2024anymaker, mao2024realcustom++}. However, none of these methods have yet explored truly zero-shot subject-driven generation.

\subsection{Zero-shot Image-to-Image Generation}
% In the realm of image-to-image (I2I) generation , some impactful works \cite{hertzp2p, avrahami2022blended, tewel2024training, kulikov2024flowedit} in zero-shot way have appeared. Techniques like \cite{hertzp2p} perform the image editing by controlling the cross-attention layers, optimizing latents \cite{nullText} obtained through DDIM inversion \cite{DDIM}, or injecting the image embedding to important attention layers \cite{stableFlow}. Approaches like \cite{styleAligned, alaluf2024cross} achieve the style transfer by sharing self-attention weights to maintain the origin layout. Blended Diffusion \cite{avrahami2022blended} spatially blend the noised version of input image with text-guided diffusion latent to accomplish image inpainting. Diffuhaul \cite{avrahami2024diffuhaul} proposes a novel interpolation between source and target images for object dragging task. Add-it \cite{addit} introduces the weighted extended-attention mechanism to add object into images. All of these I2I works have a same characteristic that the layout of synthesized image is roughly consistent with input image. Even though the subject-driven generation is also belong to I2I, it often needs the variation of layout to adapt new context, where still has not a suitable zero-shot way for this.
In the realm of image-to-image (I2I) generation, several impactful works \cite{hertzp2p, avrahami2022blended, tewel2024trainingConsiStory, kulikov2024flowedit} have embraced zero-shot approaches. Techniques such as those in \cite{hertzp2p} perform image editing by controlling cross-attention layers, optimizing latents \cite{nullText} derived via DDIM inversion \cite{DDIM}, or injecting image embeddings into key attention layers \cite{stableFlow}. Meanwhile, studies like \cite{styleAligned, alaluf2024crossSelfAttention} achieve style transfer by sharing self-attention weights to maintain the consistent layout. Blended Diffusion \cite{avrahami2022blended} spatially blends the noised version of the input with text-guided diffusion latents for inpainting, Diffuhaul \cite{avrahami2024diffuhaul} introduces novel interpolation between source and target images for object dragging, and Add-it \cite{addit} presents a weighted extended-attention mechanism to seamlessly add objects into images. All of these methods share one common characteristic: the synthesized images typically preserve a layout largely consistent with the input. In contrast, subject-driven generation, although also an I2I task, often demands layout variations to adapt to new contexts, for which an effective zero-shot solution remains elusive.

\section{Method}\label{sec:method}
We in this paper target at achieving training-free zero-shot subject-driven generation. 
Towards this, we enhance a pretrained DiT from three key perspectives.
First, we share pivotal attention from the input image during the denoising process to establish the subject layout.
Afterward, we adjust the shift in the noise scaling strength while extracting attention from the reference subject, allowing us to concentrate on fine details. 
Finally, we augment the Multimodal LLMs to incorporate essential global semantic features that may be lacking. A schematic workflow of our method is presented in Fig.~\ref{fig:overall}.

\subsection{Preliminary}
In our experiments, we adopt Flux.1 \cite{flux} as our backbone model, which builds upon the diffusion transformer (DiT) architecture \cite{peebles2023DiT}. Flux.1 trains on the latent space $z$ \cite{rombach2022LDM} of the pretrained VAE \cite{kingma2013autoVAE} model $\mathcal{E}$. Similar to SD3 \cite{sd3}, Flux.1 incorporates multi-modal self-attention blocks (MM-DiT blocks) to process sequences composed of both text and image embeddings.
%The text representations are obtained using two separate text encoders, denoted as $\tau_\theta$ \cite{clip, T5}. 
In each block, the attention operation is formulated as follows:
\begin{equation}
\small
A=\operatorname{softmax}\left(\frac{\left[Q_p, Q_{i m g}\right]\left[K_p, K_{i m g}\right]^{\top} } {\sqrt{d_k}}\right)\cdot\left[V_p, V_{i m g}\right],
\end{equation}
where $[,]$ denotes concatenation, while $Q_p$ and $Q_{img}$ represent queries from text and image embeddings, respectively, with keys $K$ and values $V$ defined similarly.

\subsection{Pivotal Attention Sharing (PAS)} \label{sec:pivotal_sharing}
To achieve the training-free customization, it is essential to integrate visual features of the reference image ($z^{ref}$) into the target image generation process. 
% A simple yet effective method \cite{styleAligned, cao2023masactrl, geyer2023tokenflow} achieves this by sharing the self-attention from $z^{ref}$ with the target image $z^{target}$, facilitating the transfer of rich spatial features.
A simple yet effective method \cite{styleAligned, cao2023masactrl, geyer2023tokenflow} achieves this by sharing the self-attention from $z^{ref}$ with the target image $z^{target}$, transferring rich spatial features.
Specifically, to transfer the self-attention from $z^{ref}$ in the DiT blocks, keys $K_{r}$ and values $V_{r}$ extracted from $z^{ref}$ are concatenated with target $K_{tgt}$ and $V_{tgt}$, while queries $Q_{p}, Q_{tgt}$ remain unchanged:
\iffalse
\begin{align}
A &= \operatorname{softmax}\left(\left[Q_p, Q_{tgt}\right]\left[K_{r}, K_p,  K_{tgt}\right]^{\top} / \sqrt{d_k}\right) \nonumber \\
&\quad \cdot \left[V_{r}, V_p, V_{tgt}\right].
\end{align}
\fi
\begin{equation}
\footnotesize
A = \operatorname{softmax}\left(\frac{\left[Q_p, Q_{tgt}\right]\left[K_{r}, K_p,  K_{tgt}\right]^{\top}} { \sqrt{d_k}}\right) \cdot \left[V_{r}, V_p, V_{tgt}\right].
\end{equation}

However, simply sharing attention significantly reduces alignment with the input prompt \cite{addit}, leading $z^{target}$ to duplicate $z^{ref}$.
Accordingly, we limit attention sharing to ten critical layers~\cite{stableFlow}, denoted as $\mathcal{V}$, highlighting the importance of these layers in influencing the generated images within the DiT model.
Additionally, since the background in $z^{ref}$ is often irrelevant or even harmful to subject customization, we extract a subject mask $m_{r}$ using an image segmentation model \cite{zheng2024birefnet} and apply masked attention sharing. The refined pivotal attention sharing (PAS) computation is defined as:
%However, the simple attention sharing drastically reduces alignment with the input prompt, causing $z^{target}$ to mimic $z^{ref}$ completely. 
%To address this issue, we constrain the communication only in ten vital layers, denoted by $\mathcal{V}$, as \cite{stableFlow} demonstrates that these layers in the DiT model significantly affect the generated images.
%
%Moreover, since the background in $z^{ref}$ is often redundant or even detrimental for subject customization, we extract a subject mask $m_{r}$ using an image segmentation model \cite{zheng2024birefnet} and perform masked attention sharing. The updated, pivotal attention sharing (PAS) computation is defined as:

\begin{align}
A_l= 
\begin{cases} 
\operatorname{softmax}\left(\frac{Q \cdot K'^{\top}}{\sqrt{d_k}}\right) \cdot V' & \text{if } l \in \mathcal{V} \\ 
\operatorname{softmax}\left(\frac{Q \cdot K^{\top}}{\sqrt{d_k}}\right) \cdot V & \text{otherwise}
\end{cases},
\label{eq:sharing}
\end{align}
\text{where:}
% \begin{align*}
% &Q=\left[Q_p, Q_{tgt}\right],K=\left[K_p,  K_{tgt}\right], \\
% &V=\left[V_p, V_{tgt}\right], V'=\left[V_{r} \odot m_{r}, V_p, V_{tgt}\right], \\
% &K'=\left[\lambda_{r} \cdot K_{r} \odot m_{r}, \lambda_{p} \cdot K_p,  K_{tgt}\right].
% \end{align*}
\begin{align*}
&Q=\left[Q_p, Q_{tgt}\right],K=\left[K_p,  K_{tgt}\right], V=\left[V_p, V_{tgt}\right], \\
&K'=\left[\lambda_{r} \cdot K_{r} \odot m_{r}, \lambda_{p} \cdot K_p,  K_{tgt}\right], \\
&V'=\left[V_{r} \odot m_{r}, V_p, V_{tgt}\right].
\end{align*}
Since $K_{r}$ and $K_p$ critically govern the subject consistency and text alignment, we employ scalars $\lambda_{r}$ and $\lambda_{p}$ to control the relative influence of $z^{ref}$ and the target prompt.

\myparagraph{Attention of the reference image.} The shared attentions are obtained by denoising intermediate noisy samples of the reference image $z^{ref}$ at all timesteps, referred to as the diffusion trajectory $z_T, z_{T-1},...,z_0$. Consequently, accurate recovery of the diffusion trajectory is essential.
Image inversion techniques \cite{rout2024semantic, DDIM} are typically employed to obtain these intermediate samples. However, such methods often fail \cite{nullText} or produce erroneous trajectories \cite{huberman2024editFriendly}. Instead, we inject random noise $\epsilon$ into $z^{ref}$ via a rectified flow forward process \cite{lipmanflow, liu2023flow, albergobuilding} to generate the trajectory:
\begin{equation}
    z_{t} = (1-\sigma_t)z_0 + \sigma_t \epsilon,
    \label{eq:forward}
\end{equation}
where $\sigma_t$ represents the strength of noise scaling. Although these noisy samples are derived from random noise, the resulting trajectory remains valid. Inaccuracies in the attention computed at high timesteps are progressively corrected as the noise diminishes, since $\sigma_0=0$ ensures $z_0 = z^{ref}$. Thus, by denoising these samples, we reliably obtain the desired attention features from the reference image.

\begin{figure}[!t]
  \centering
  \includegraphics[width=\linewidth]{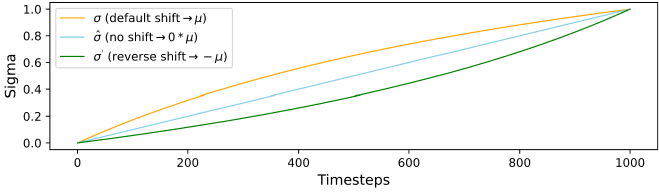}
  \caption{\textbf{The noise scaling $\sigma$ under different shift directions across all timesteps at a target resolution of $\textbf{512} \times \textbf{512}$.} 
  %`default' corresponds to the default shift value $\mu$ in the dynamic shifting of Flux.1. `w/o shift' indicates a shift value of $0 * \mu$, while `reverse' uses a shift value of $-\mu$.
  }
  %  As shown in the illustration, the default configuration emphasizes higher noise strengths, whereas modifying the shift direction shifts this focus.
  \label{fig:sigmas}
  \vspace{-15pt}
\end{figure}

\begin{figure*}[!t]
  \centering
  \includegraphics[width=\linewidth]{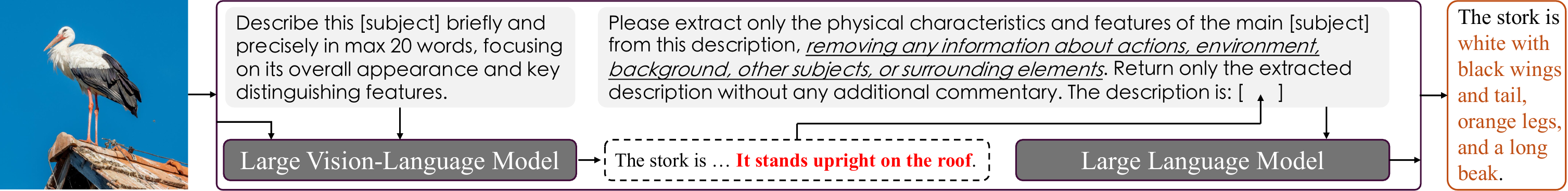}
  \caption{
  %\textbf{Subject caption by multimodal LLMs.} To get the appropriate caption of the reference subject, we first leverage a large vision-language model to get complete but redundant caption, then use a large language model to filter harmful information (highlight in red of the figure) for subject-driven generation in new context. 
  \textbf{Illustration of the subject caption generation process with Multimodal LLMs.} 
  %To derive an appropriate caption for the reference subject, we first employ a large vision-language model to produce a comprehensive yet redundant caption. Next, a large language model filters out harmful or irrelevant details (highlighted in \textcolor{red}{red} in the figure), ensuring the caption is suitable for subject-driven generation in a new context.
  }
  \label{fig:mllm}
  \vspace{-15pt}
\end{figure*}

% \subsection{Time Shifting for Fine Details}
% \subsection{Adjustment of Time Shifting} \label{sec:adjustment_shift}
\subsection{Adjustment of Noise Shifting (ANS)} \label{sec:adjustment_shift}

% Given that we perform attention sharing on only ten vital layers, some subject details are inevitably lost. To mitigate this, we analyze the dynamic timestep shifting in Flux.1 and propose an adjusted version of \cref{eq:forward} that preserves finer details.
As we restrict attention sharing to ten vital layers, some subject details are inevitably lost. To address this, we analyze dynamic shifting in Flux.1 and propose an adjusted version of \cref{eq:forward} to preserve finer details.
%Given that we perform attention sharing on only ten vital layers, some subject details are inevitably lost. To mitigate this, we analyze the dynamic shifting in Flux.1 and propose an adjusted version of \cref{eq:forward} that preserves finer details.
%
% The mentioned dynamically shifted timestep is computed as follows:
% \begin{equation}
%     t_{shift}=\sigma_t \cdot T_{train}, \sigma_t= \frac{e^\mu}{e^\mu+\frac{1}{t}-1},\mu = L_x\cdot m+b,
% \end{equation}
The mentioned dynamically shifted noise scaling $\sigma_t$ is computed as follows:
\begin{equation}
    \sigma_t= \frac{e^\mu}{e^\mu+\frac{1}{t}-1}, \mu = L_x\cdot m+b,
    \label{eq:ds}
\end{equation}
where $t$ represents the current timestep, $L_x$ is the latent sequence length of the target image computed by the VAE's scale factor and image resolution, $m$ and $b$ are fixed constants, and $\mu$ denotes the dynamic shift, which increases with image resolution. Noise levels under this dynamic shifting (derived from \cref{eq:ds}) are consistently higher than those in the ``no shifting'' setting (i.e., $\sigma_t\ge \hat{\sigma_t}$ as shown in \cref{fig:sigmas}), guiding the model to focus on noisier samples via \cref{eq:forward}, which is suitable for higher-resolution images requiring greater signal impairment \cite{sd3}.

However, to extract finer details from the reference image $z^{ref}$, we emphasize lower noise levels for $z^{ref}$. To achieve this, we reverse the shifting direction ($\sigma^{'}$ in Fig. \cref{fig:sigmas} ) when computing attentions for $z^{ref}$. The modified noise scaling at timestep $t$ is defined as $\sigma_t^{'} = \frac{e^{-\mu}}{e^{-\mu} + \frac{1}{t} - 1}$, resulting in a new diffusion trajectory: $z_t = (1-\sigma_t^{'})z^{ref} + \sigma_t^{'} \epsilon$. This adjustment of noise shifting (ANS) ensures that attentions prioritize less noisy, subject-specific content from $z^{ref}$ (see \cref{fig:sigmas}), enabling finer detail transfer to the target image during attention sharing. Ablation studies in \cref{sec:ablation} further evaluate different shift directions to identify the optimal configuration.

\subsection{Semantic Features Compensation (SFC)} \label{sec:sfc}
In addition to fine details, semantic features, such as color, can be compromised due to the limited extent of attention sharing. To address this, we use advanced Multimodal LLMs \cite{qwen2.5, Qwen2VL} to generate a concise, subject-specific caption (see \cref{fig:mllm}).
First, the reference image is input into a large vision-language model (LVLM) \cite{Qwen2VL}, leveraging its strong visual understanding to generate captions. The output is constrained to 20 tokens, focusing on key attributes to avoid irrelevant details. As demonstrated in \cref{sec:ablation}, a streamlined caption performs better than a detailed one. However, LVLMs may still include unrelated information, such as background or actions (highlighted in \textcolor{red}{red} in \cref{fig:mllm}), which can mislead subsequent image generation.
To resolve this, we use a large language model (LLM) \cite{qwen2.5} to filter out irrelevant details, capitalizing on its robust natural language processing capabilities. This process produces a refined caption that emphasizes essential subject attributes. The caption is then combined with the original prompt to address semantic feature deficiencies, ensuring a more accurate and comprehensive subject representation.

\section{Experiments}
\subsection{Experimental Settings} \label{sec:settings}
% \myparagraph{Datasets.} 
% \myparagraph{Implementation details.} We exploit the pretrained Flux.1-dev \cite{flux} as the base model. We perform the inference with 30 steps, a guidance scale of 3.5, and $512 \times 512$ resolution. Hyper parameters including $\lambda_r$ and $\lambda_p$ are empirically set to $1.1$ and $1.1$. The used advancing segmentation model, large vision-language model, and large language model are BirefNet \cite{zheng2024birefnet}, Qwen2-VL-7B-Instruct \cite{Qwen2VL}, and Qwen2.5-7B-Instruct \cite{qwen2.5}, respectively. Besides, due to the rapid growing of multimodal LLMs, the performance of our approach will be enhanced by using stronger new model.

\myparagraph{Implementation details.} We adopt the pretrained Flux.1-dev \cite{flux} as our base model. Inference is performed using 30 steps, a guidance scale of 3.5, and a resolution of $512 \times 512$. The hyperparameters, including $\lambda_r$ and $\lambda_p$, are empirically set to $1.1$. For advancing segmentation, large vision-language, and large language models, we utilize BirefNet \cite{zheng2024birefnet}, Qwen2-VL-7B-Instruct \cite{Qwen2VL}, and Qwen2.5-7B-Instruct \cite{qwen2.5}, respectively. Furthermore, as Multimodal LLMs continue to advance rapidly, the performance of our approach is expected to improve with the integration of stronger models.

% \myparagraph{Evaluation metrics.} We use the DreamBench++ \cite{peng2024dreambench++} as the evaluate benchmark, which is 5× more than commonly used DreamBench \cite{ruiz2023dreambooth}. It collects 150 images and 1,350 prompts, covering various categories from animals and styles that are relatively simple to more challenging human subjects, objects. Here we evaluate results on three classes including animal, human and object. Besides, the style transfer would be viewed as the extend of our approach. This diverse and rich evaluation data is necessary to avoid biased evaluation. For the quantitative indicators, we evaluate results on subject similarity and text controllability. For the former, we use the CLIP-I and DINO \cite{caron2021emergingDINO} scores, computing the average pairwise cosine similarity of CLIP or DINO embeddings of generated and given subjects, to measure subject consistency. Notably, we use the segmentation model SAM \cite{sam} to get segmented subjects, like to \cite{mao2024realcustom++}. For the latter, we use the CLIP-T score, calculating the cosine similarity between the prompt and the image CLIP embeddings, to assess the consistency between image and prompt. For the evaluation suite we generate four images per subject and per prompt.

\myparagraph{Evaluation metrics.} We evaluate our approach using the DreamBench++ benchmark \cite{peng2024dreambench++}, which is five times larger than the commonly used DreamBench \cite{ruiz2023dreambooth}. 
% DreamBench++ comprises 150 images and 1,350 prompts, covering a broad range of categories—from relatively simple subjects (e.g., animals and basic styles) to more challenging subjects (e.g., human figures and diverse objects). In our evaluation, we focus on three classes: animal, human, and object, while also considering style transfer as an extension of our approach. This diverse dataset helps mitigate evaluation bias.
%
For quantitative assessment, we use two primary metrics. First,     the subject similarity is evaluated using CLIP-I and DINO \cite{caron2021emergingDINO} scores by computing the average pairwise cosine similarity between the embeddings of the generated subjects and the corresponding reference subjects. To ensure accurate comparison, we employ the segmentation model SAM \cite{sam} to isolate the subject regions, following the methodology of \cite{mao2024realcustom++}. Second, text controllability is assessed with the CLIP-T score, which measures the cosine similarity between the prompt and the image CLIP embeddings, thereby gauging the consistency between the generated image and the input prompt. For each subject and prompt pair, four images are generated to form the evaluation suite.

% single subject personalization?  diverse subjects ?
% \myparagraph{Compared methods.} We conduct the comparison with two main streams across different base models: 1) Optimization-based methods that require retraining for each new subject, including Textual Inversion (TI) \cite{textualInversion}, DreamBooth \cite{ruiz2023dreambooth}, and DreamBooth LoRA (DreamBooth-L) \cite{ruiz2023dreambooth, hulora}; 2) Optimization-free customization methods that trains on large-scale datasets, including BLIP-Diffusion \cite{li2024blipDiffusion}, Emu2 \cite{sun2024generativeEmu2}, IP-Adapter-Plus \cite{ye2023ipa}, IP-Adapter \cite{ye2023ipa} (two versions performed on SDXL \cite{podellsdxl} and Flux.1, respectively), MS-Diffusion \cite{msDiffusion}, and Qwen2VL-Flux \cite{erwold-2024-qwen2vl-flux}. Some results of these methods are from the Dreambench++. Details of methods are in the supp.

\myparagraph{Compared methods.} 
We compare our approach with two main streams of customization methods across different base models: 1) Optimization-based methods that require retraining for each new subject, including Textual Inversion (TI) \cite{textualInversion}, DreamBooth \cite{ruiz2023dreambooth}, and DreamBooth LoRA (DreamBooth-L) \cite{ruiz2023dreambooth, hulora}; 2) Optimization-free customization methods trained on large-scale datasets, such as BLIP-Diffusion \cite{li2024blipDiffusion}, Emu2 \cite{sun2024generativeEmu2}, IP-Adapter-Plus \cite{ye2023ipa}, IP-Adapter \cite{ye2023ipa} (implemented on both SDXL \cite{podellsdxl} and Flux.1), MS-Diffusion \cite{msDiffusion}, Qwen2VL-Flux \cite{erwold-2024-qwen2vl-flux} and OminiControl \cite{tan2024ominicontrol}.
Some results for these methods are obtained from DreamBench++ implementations, and further details are provided in the supplementary material.

\subsection{Comparison Results} 
\begin{figure*}[!t]
  \centering
  \includegraphics[width=\linewidth]{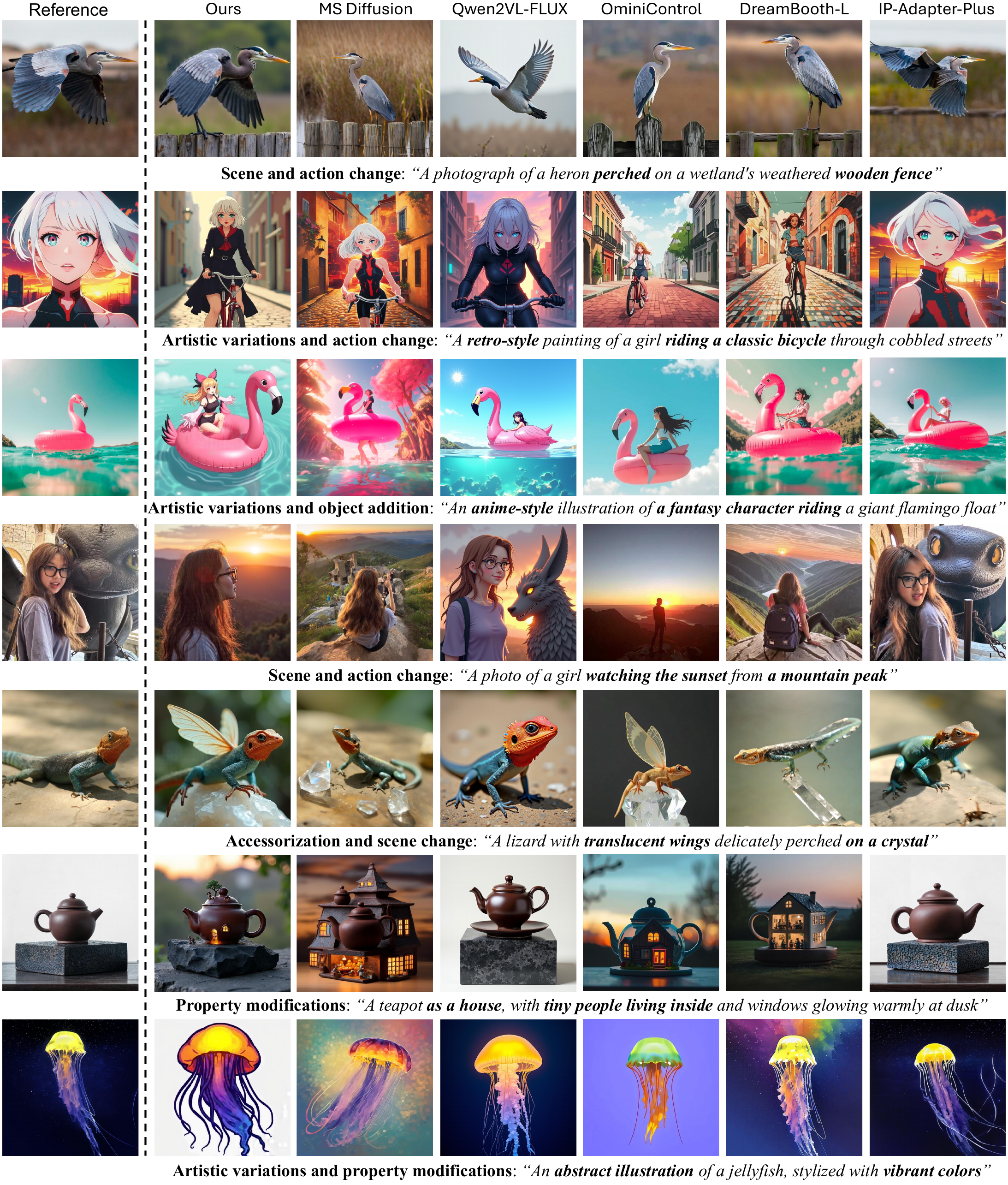}
  \caption{\textbf{Qualitative evaluation results.} Comparison across various subjects and contexts reveals: OminiControl and DreamBooth-L lack subject fidelity; IP-Adapter-Plus and Qwen2VL-Flux fail at text alignment; MS-Diffusion generates background artifacts (rows 1 and 4). In contrast, our method successfully balances subject fidelity with prompt adherence while generating high-quality images.}
  \label{fig:main_comparison}
  
\end{figure*}
% \myparagraph{Quantitative comparisons.} \cref{table:main_comparsion} shows the results averaged on three classes, and the version including each class in the supp. 
% %
% As shown, thanks to robust feature extractors trained on large-scale datasets, optimization-free methods clearly surpass the optimization-based methods on the subject similarity. But optimization-based methods is still have the advantage on the text controllability, as they usually make smaller changes on the generation distribution of the base model, such as the DreamBooth-L. 
% %
% Furthermore, we observe that the base model stronger, the overall generation results better, as the IP-Adapter (Flux.1) scores higher than IP-Adapter (SDXL) on two indicates. 
% %
% Different with others, IP-Adapter-Plus expresses highest subject similarity with the large sacrifice of text controllability. 
% %
% The MS-Diffusion looks like that it has the best trade-off between the two metrics. However, it shows a big problem in the qualitative results (shown in the below section).
% %
% Even though our method is not need to optimize embeddings or train encoders, \textit{FreeCus} surpass most methods on subject similarity and has a good text controllability. This is because that our training-free paradigm makes full use of the robust pretrained features (similar to the advantage of optimization-free methods) while carefully change the output distribution of the base model (similar to the advantage of optimization-free methods) due to the proposed pivotal attention sharing. 

\myparagraph{Quantitative comparisons.} 
\cref{table:main_comparsion} presents averaged results across three classes (animal, human, object), with per-class details in the supplementary material.
As shown, optimization-free methods, benefiting from robust feature extractors trained on large datasets, clearly outperform optimization-based methods (marked with \(^{\dagger}\)) in subject similarity (CLIP-I and DINO scores), while the latter maintain better text controllability (CLIP-T scores) by making smaller adjustments to the base model's output distribution (e.g., DreamBooth-L and OminiControl).
Furthermore, we observe that stronger base models generally perform better, as evidenced by IP-Adapter (Flux.1) surpassing IP-Adapter (SDXL) on two metrics. 
While IP-Adapter-Plus achieves the highest subject similarity, it significantly compromises text controllability. MS-Diffusion appears to offer the best trade-off across metrics, though its qualitative performance has notable limitations (discussed later).

Our method, without requiring embedding optimization or encoder training, surpasses most competitors in subject similarity while maintaining good text controllability. This is attributed to our training-free paradigm, which fully leverages robust pretrained features (similar to optimization-free methods) and carefully adjusts the output distribution of the base model via the proposed strategies.

\myparagraph{Qualitative comparisons.} Considering page constraints, we present qualitative results from five controversial methods listed in \cref{table:main_comparsion}. 
This comparison covers various subject categories (animal, object, human, anime character) and functionalities (scene changes, object addition, artistic variations, property modifications, accessorization, and action changes), as illustrated in Figure~\ref{fig:main_comparison}.
OminiControl and DreamBooth-L exhibit strong instruction-following capabilities but compromise subject consistency. While IP-Adapter-Plus achieves high subject fidelity, it essentially sacrifices text controllability. 
Qwen2VL-Flux shows similar limitations in disentangling multimodal embeddings due to its text embedding replacement strategy via Qwen2-VL.
Although MS-Diffusion leads in quantitative metrics, it produces noticeable artifacts in synthesized backgrounds (see rows 1 and 4 in \cref{fig:main_comparison}).
% MS-Diffusion has the best, comprehensive scores in \cref{table:main_comparsion}, but shows severe chaos in the background, shown in the first and forth row of \cref{fig:main_comparison}. 
%
In contrast, our method achieves high subject fidelity while enabling diverse contextual adaptations, demonstrating its potential to extend to more fantastical subject-driven generation. 

\begin{table}[]
\centering
\resizebox{\linewidth}{!}{
\begin{tabular}{lcccc}
\hline
Method                      & BaseModel                  & CLIP-T ↑                     & CLIP-I ↑                     & DINO ↑                       \\ \hline
Textual Inversion$^\dagger$ & SD v1.5                    & 0.298                        & 0.713                        & 0.430                        \\
DreamBooth$^\dagger$        & SD v1.5                    & {\color[HTML]{3166FF} 0.322} & 0.716                        & 0.505                        \\
DreamBooth-L$^\dagger$      & SDXL v1.0                  & {\color[HTML]{3166FF} 0.341} & 0.751                        & 0.547                        \\
BLIP-Diffusion              & SD v1.5                    & 0.276                        & 0.815                        & 0.639                        \\
Emu2                        & SDXL v1.0                  & 0.305                        & 0.763                        & 0.529                        \\
IP-Adapter                  & SDXL v1.0                  & 0.305                        & 0.845                        & 0.621                        \\
IP-Adapter-Plus             & SDXL v1.0                  & 0.271                        & {\color[HTML]{3166FF} 0.916} & {\color[HTML]{3166FF} 0.807} \\
MS-Diffusion                & SDXL v1.0                  & {\color[HTML]{3166FF} 0.336} & {\color[HTML]{3166FF} 0.873} & {\color[HTML]{3166FF} 0.729} \\
Qwen2VL-Flux                & FLUX.1                     & 0.267                        & 0.841                        & 0.664                        \\
IP-Adapter                  & FLUX.1                     & 0.314                        & 0.840                        & 0.638                        \\
OminiControl                & FLUX.1 & {\color[HTML]{3166FF} 0.330} & 0.797                        & 0.570                        \\ \hline
Ours                        & FLUX.1                     & \textbf{0.308}               & \textbf{0.853}               & \textbf{0.696}               \\
w/o PAS                     & FLUX.1                     & 0.327                        & 0.810                        & 0.590                        \\
w/o ANS                     & FLUX.1                     & 0.324                        & 0.829                        & 0.624                        \\
w/o SFC                     & FLUX.1                     & 0.322                        & 0.822                        & 0.633                        \\ \hline

\end{tabular}
}
% \caption{Quantitative evaluation results. 1) Even though the no need of any training resource, our training-free approach outperforms most training-oriented on subject similarity (CLIP-I and DINO indicators), while maintaining good text alignment (CLIP-T indicator). {\color[HTML]{3166FF} Blue} represents the better value by compared methods, and $^\dagger$ represents the optimization-based methods. 2) Ablation study results shows each component has an important impact on the subject similarity. {\color[HTML]{CE6301} Brown} represents the better value by ablation methods.}
% \caption{\textbf{Quantitative evaluation results.} (1) Despite without any training resources, our zero-shot approach outperforms most training-oriented methods in subject similarity (CLIP-I and DINO scores), while maintaining strong text alignment (CLIP-T score). {\color[HTML]{3166FF} Blue} represents higher scores than our approach by compared methods, and \(^{\dagger}\) marks optimization-based methods. (2) Ablation study results demonstrate that each component significantly impacts subject similarity.}
\caption{\textbf{Quantitative evaluation results.} {\color[HTML]{3166FF} Blue} indicates scores higher than ours, and \(^{\dagger}\) denotes optimization-based methods. 
%(1) Our approach, requiring no training resources, surpasses most training-based methods in subject similarity (CLIP-I and DINO) while preserving text alignment (CLIP-T). (2) Ablation studies confirm each component's significant contribution to subject similarity.
}
\vspace{-15pt}
\label{table:main_comparsion}
\end{table}

\begin{figure}[!t]
  \centering
  \includegraphics[width=\linewidth]{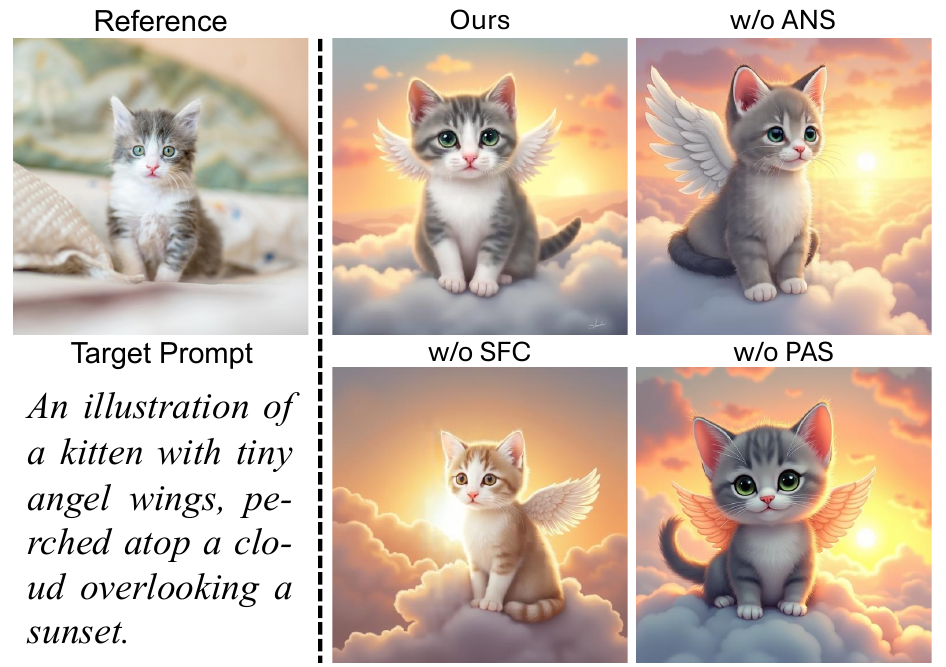}
  \caption{\textbf{Impact visualization of each proposed component.} %Compared to our full method, the absence of time shifting adjustment (w/o ATS) results in the loss of detailed features, such as facial and leg textures. Removing semantic feature compensation (w/o SFC) leads to inconsistent semantic information, exemplified by discrepancies in body and eye colors. Excluding pivotal attention sharing (w/o PAT) causes significant differences between the input and output.
  }
  % \Description{A woman and a girl in white dresses sit in an open car.}
  \label{fig:ablation}
  \vspace{-17pt}
\end{figure}

\subsection{Ablation Studies} \label{sec:ablation}
% \myparagraph{Ablation studies on each component.}
% First, we conduct ablation studies on each component with quantitative and qualitative analysis. 
% %
% As shown in \cref{table:main_comparsion}, removing any individual module leads to the severe decline on the subject similarity. Through \cref{fig:ablation}, we can intuitively observe the reasons. Compared to our full method, the model would lose the texture and other details without adjustment of shift type (shown in the w/o ATS in \cref{fig:ablation}), e.g. the cat's hair texture especially in its leg and face. This is due to the origin time shifting that favors higher noise strength, analyzed in \cref{sec:adjustment_shift}.
% %
% Throwing away the semantic caption of the subject, the model is prone to generating content different with input in semantic features, like inconsistent color in the body and eyes with input shown in the w/o SFC in \cref{fig:ablation}. 
% %
% Canceling the pivotal attention sharing leads to the most inconsistent results on quantitative and qualitative, as the lowest CLIP-I and DINO scores in \cref{table:ablation} and only rough features similar to the specific cat in \cref{fig:ablation}. 

\myparagraph{Ablation studies on each component.}
We conduct ablation studies with both quantitative and qualitative analysis to evaluate the contribution of each component. 
As shown in \cref{table:main_comparsion}, removing any individual module significantly reduces subject similarity. Through visual inspection of \cref{fig:ablation}, we can identify specific degradations. Without the adjustment of shift type (w/o ANS), the model fails to preserve fine-grained textures and details, particularly evident in the cat's facial features and leg fur. This occurs because the default dynamic shifting mechanism prioritizes higher noise strength, overwhelming subject details, as discussed in \cref{sec:adjustment_shift}.
Removing the semantic caption (w/o SFC) leads to inconsistent semantic features, such as mismatched body and eye coloration.
The most significant performance drop occurs when pivotal attention sharing is removed (w/o PAS), resulting in the lowest CLIP-I and DINO scores in \cref{table:main_comparsion}. Visually, this ablation retains only rough features of the reference cat, as illustrated in \cref{fig:ablation}. 
% ablation for the important layers
% Moreover, ablation concerning vital layer selection is presented in the supplementary materials.
Moreover, ablation of vital layer selection is in supplementary.

\iffalse
As shown in \cref{table:main_comparsion}, removing any individual module significantly reduces subject similarity. Through visual inspection of \cref{fig:ablation}, we can identify specific degradations. Without the adjustment of shift type (w/o ANS), the model fails to preserve fine-grained textures and details, particularly evident in the cat's facial features and leg fur. This deterioration occurs because the original dynamic shifting mechanism favors higher noise strength, as analyzed in \cref{sec:adjustment_shift}.
%
When semantic caption is removed (w/o SFC), the model generates content with inconsistent semantic features compared to the input, such as discrepancies in body and eye coloration as demonstrated in \cref{fig:ablation}.
%
Among all ablations, removing pivotal attention sharing (w/o PAS) produces the most substantial performance degradation, reflected in the lowest CLIP-I and DINO scores in \cref{table:main_comparsion}. Visually, this ablation preserves only rough features of the reference cat, as evidenced in \cref{fig:ablation}.

\fi
% three aspects, scale of different attn, shift, and semantic prompt 
\myparagraph{Hyperparameter analysis.}
For pivotal attention sharing, we examine the impact of the reference image and target text, controlled by the hyperparameters $\lambda_r$ and $\lambda_p$ in \cref{eq:sharing}, which are set to the same value for simplicity. Ablative details are presented below:
%For pivotal attention sharing, we focus on the influence of the reference image and target text, represented by the hyperparameters $\lambda_r$ and $\lambda_p$ in \cref{eq:sharing}, which we empirically set to the same value. The ablation results are presented below:
\begin{table}[h]
\centering
\vspace{-10pt}
\resizebox{0.7\linewidth}{!}{
    \begin{tabular}{l|ccc}
    $\lambda_p$,   $\lambda_r$ & CLIP-T ↑       & CLIP-I ↑       & DINO ↑         \\ \hline
    1.00                       & 0.321          & 0.827          & 0.626          \\
    1.05                       & 0.315          & 0.838          & 0.656          \\
    \textbf{1.10}              & \textbf{0.308} & \textbf{0.853} & \textbf{0.696} \\
    1.15                       & 0.305          & 0.861          & 0.706         
    \end{tabular}
}

\vspace{-13pt}
\end{table}

The above results reveal a trade-off: increasing $\lambda_r$ and $\lambda_p$ improves subject similarity (higher CLIP-I and DINO) but slightly reduces text alignment (lower CLIP-T). We select $\lambda_p = \lambda_r = 1.10$ as the optimal configuration, balancing subject fidelity and text controllability, as further increases yield diminishing returns in subject consistency.

%The results demonstrate a clear trade-off: increasing $\lambda_r$ and $\lambda_p$ enhances subject similarity (higher CLIP-I and DINO scores) while slightly reducing text alignment capabilities (lower CLIP-T scores). We select $\lambda_p = \lambda_r = 1.10$ as the optimal configuration, as it achieves an effective balance between subject fidelity and text controllability, with further increases yielding diminishing returns in subject consistency.
% As $\lambda_r$ and $\lambda_p$ increase, subject similarity improves, while text-following ability slightly declines. We ultimately set $\lambda_p = \lambda_r = 1.10$, balancing the diminishing improvement in subject consistency with the need for strong text controllability.
%For the pivotal attention sharing, we mainly focus on the influence of reference image and target text, that is the hyper parameters $\lambda_r$ and $\lambda_p$ in \cref{eq:sharing}, and empirical make them same value. 

%As $\lambda_r$ and $\lambda_p$ increases, the scores of subject similarity ascend while text following ability declines, shown in the left of \cref{table:ablation}. We final set $\lambda_p=\lambda_r=1.10$, considering that the improvement on subject consistency is slow after it and the need of a good text controllability.

\myparagraph{Shift type analysis.} 
A similar trade-off phenomenon appears in the time shifting type, as analyzed in \cref{sec:adjustment_shift}. The quantitative results are presented below:
\begin{table}[h]
\centering
\vspace{-10pt}
\resizebox{0.7\linewidth}{!}{
    \begin{tabular}{l|ccc}
    Shift type   & CLIP-T ↑       & CLIP-I ↑       & DINO ↑         \\ \hline
    $\mu$ * 0    & 0.320          & 0.836          & 0.648          \\
    $\mu$ * -0.5 & 0.315          & 0.845          & 0.670          \\
    $\mu$ * -1.0 & \textbf{0.308} & \textbf{0.853} & \textbf{0.696} \\
    $\mu$ * -2.0 & 0.296          & 0.857          & 0.698       
    \end{tabular}
}
\vspace{-13pt}
\end{table}

As shown above, increasing the negative shift magnitude ($-\mu$) enhances subject similarity but reduces text instruction adherence.
 Paralleling our reasoning for the pivotal attention sharing parameters, $\mu * -1.0$ is selected as optimal.

\myparagraph{Designs for captions.} We explore four strategies for subject caption generation to identify the most suitable way:
1) Using a large vision-language model (LVLM) to generate concise, general subject descriptions (+ LVLM);
2) Employing LVLM with specialized prompts to create detailed subject descriptions (+ detailed LVLM, prompts provided in supplementary materials);
3) Applying a large language model (LLM) to filter the general LVLM outputs, thereby eliminating harmful annotations (+ filtered LVLM), as detailed in \cref{sec:sfc};
4) Implementing LLM filtering on detailed LVLM descriptions (+ detailed, filtered LVLM).
% Besides, we also attempt to design different subject caption for the best generation quality: 1) using LVLM to obtain overall, roughly subject caption, i.e. `+ LVLMs, 2) using LVLM to obtain detailed subject caption with different prompt shown in the supp, i.e. `+ detailed LVLM', 3) using the LLM to filter results from `+ LVLM' to exclude noisy annotations, 4) using the LLM to filter results from `+ detailed, filtered LVLM'. 
\begin{table}[h]
\centering
\vspace{-9pt}
\resizebox{0.9\linewidth}{!}{
    \begin{tabular}{l|ccc}
Caption                   & CLIP-T ↑       & CLIP-I ↑       & DINO ↑         \\ \hline
+ LVLM                    & 0.303          & 0.860          & 0.709          \\
+ detailed LVLM           & 0.303          & 0.856          & 0.700          \\
+ filtered LVLM           & \textbf{0.308} & \textbf{0.853} & \textbf{0.696} \\
+ detailed, filtered LVLM & 0.308          & 0.848          & 0.682         
\end{tabular}
}

\vspace{-12pt}
\end{table}

Our results demonstrate that overly detailed captions adversely affect subject-driven generation by introducing contextual constraints that limit adaptability. The quantitative metrics show that filtered LVLM captions strike the optimal balance between text alignment (CLIP-T) and subject fidelity (CLIP-I and DINO). While unfiltered LVLM captions yield marginally higher subject similarity scores, the filtered approach provides superior text controllability.

\subsection{Applications}

\begin{figure}[!t]
  \centering
  \vspace{7pt}
  \includegraphics[width=\linewidth]{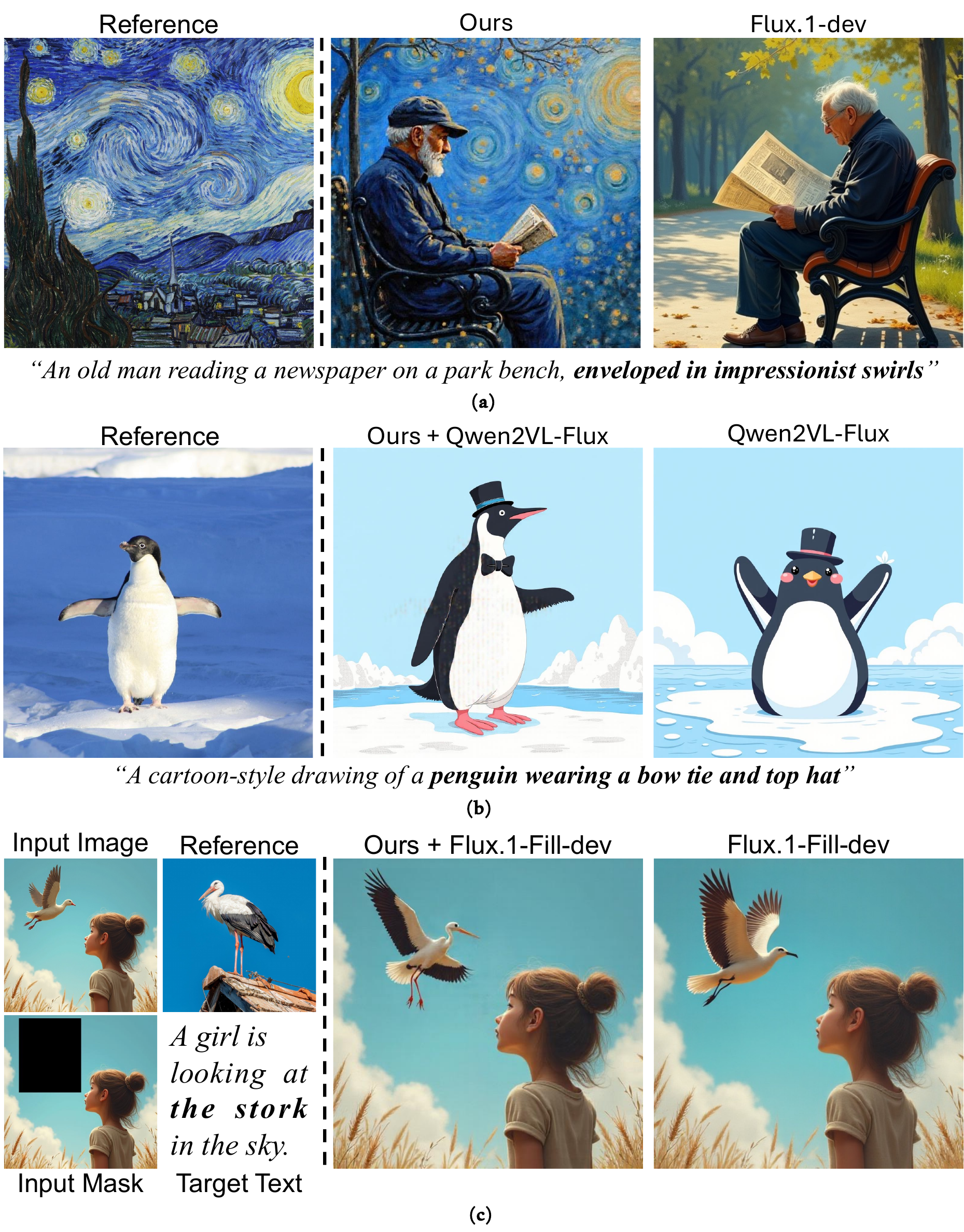}
  \caption{\textbf{Extending to more applications.} (a) Applying our method to the style transfer task; (b) Compatibility with other methods; (c) Integration with inpainting pipeline.}
  % \Description{A woman and a girl in white dresses sit in an open car.}
  \label{fig:applications}
  \vspace{-7pt}
\end{figure}

\myparagraph{Style-aligned image generation.}
Style can be conceptualized as an abstract subject permeating the entire image. 
While the base model fails to interpret specific styles from textual descriptions, our approach successfully integrates these styles into the generated images, as demonstrated in \cref{fig:applications}(a). This adaptation requires only a modification to the prompt for semantic subject caption generation (details provided in supplementary materials).

\myparagraph{Compatibility with other methods.}
% The zero-shot nature of our approach enables seamless integration with other DiT architecture-based methods, enhancing their performance. For instance, applying our method to Qwen2VL-Flux yields superior results compared to the original implementation. As illustrated in \cref{fig:applications}(b), the penguin generated by ``ours + Qwen2VL-Flux'' exhibits greater fidelity to the input image and correctly includes the bow tie, a detail absent in the standard Qwen2VL-Flux output. 
% More quantitative improvements are documented in the supplementary materials.
The zero-shot nature of our approach enables seamless integration with other DiT-based methods, enhancing their performance. For instance, applying it to Qwen2VL-Flux outperforms the original model. As illustrated in \cref{fig:applications}(b), the penguin generated by ``ours + Qwen2VL-Flux'' exhibits greater fidelity to the input image and correctly includes the bow tie, a detail absent in the standard Qwen2VL-Flux output. 
Quantitative improvements are detailed in the supplementary.

\myparagraph{Subject-driven inpainting.}
Our method naturally extends to personalized image inpainting tasks using the Flux.1-Fill-dev model. Since this model requires a mask as input, we initially use a completely black mask to achieve perfect reconstruction of the reference image. This process yields accurate shared attention weights, as described in \cref{sec:pivotal_sharing}. Subsequently, we apply our paradigm during the inpainting process. As shown in \cref{fig:applications}(c), our approach seamlessly integrates the reference subject into the masked region while preserving the integrity of the surrounding image. Additionally, our method can be applied to the Flux.1-Depth-dev model to control the structural properties of the target image (visual illustrations provided in supplementary materials).

\section{Conclusions and Limitations} 
We propose \textit{FreeCus} for truly training-free subject-driven generation through three novel strategies on pretrained diffusion transformers. First, we introduce pivotal attention sharing to effectively mimic the subject's layout while maintaining strong editability. Second, we revise DiT's dynamic shifting mechanism to enhance detail preservation in the shared attention maps. Third, we leverage Multimodal LLMs to generate subject-appropriate captions that compensate for potential semantic feature deficiencies. Our extensive experiments demonstrate that \textit{FreeCus}, despite operating in a zero-shot manner, achieves performance comparable to or exceeding state-of-the-art methods trained on large-scale datasets. We further validate our method's versatility through diverse application scenarios.

\myparagraph{Limitations.} 
 Our approach faces two primary challenges. First, the attention sharing mechanism occasionally introduces artifacts with outlines resembling the reference subject. While we attempted to mitigate this by shifting position indices of shared attention \cite{tan2024ominicontrol}, this solution reduced subject similarity. This highlights the ongoing challenge of developing more flexible methods for reference feature mapping. Second, subject captions from Multimodal LLMs aren’t fully accurate yet. We anticipate that rapid advancements in multimodal language modeling will address this limitation in the near future.
 
%Our approach faces two primary challenges. First, the attention sharing mechanism occasionally introduces artifacts with outlines resembling the reference subject. While we attempted to mitigate this by shifting position indices of shared attention \cite{tan2024ominicontrol}, this solution reduced subject similarity. This highlights the ongoing challenge of developing more flexible methods for reference feature mapping. Second, subject captions generated by multimodal LLMs still are not entirely accurate. We anticipate that rapid advancements in multimodal language modeling will address this limitation in the near future.

\newpage
\myparagraph{Acknowledgment.} This work is supported by Natural Science Foundationof China under Grant No. 62476087, Shanghai Municipal Education Commission's Initiative on Artificial Intelligence-Driven Reform of Scientific Research Paradigms and Empowerment of Discipline Leapfrogging,  Natural Science Foundation of China under Grant No. 62201341, National Key Research and Development Program of China under Grant No. 2022YFB3203500.
{
    \small
    \bibliographystyle{ieeenat_fullname}
    \bibliography{main}
}

\newpage
\clearpage
\setcounter{page}{1}
\maketitlesupplementary

\section{Experiments}

\myparagraph{Ablations on vital layer selection.} We investigate: Does the benefit arise from simply reducing layers or specifically using vital layers? Do non-vital layers impact generation? Does attention-dropout \cite{tewel2024trainingConsiStory} suffice? 

\begin{figure}[h]
  \centering
  \vspace{-3pt}
  \includegraphics[width=\linewidth]{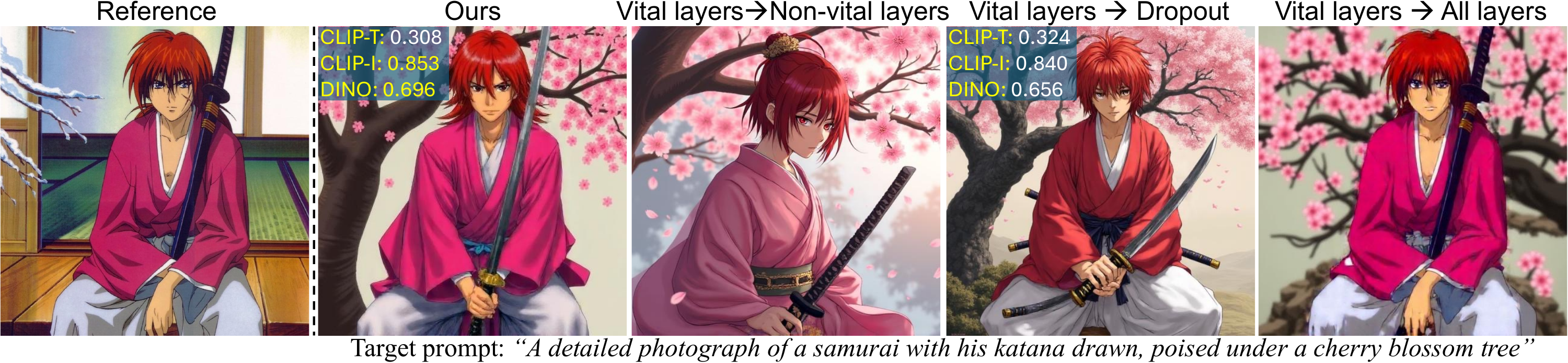}
  \vspace{-7pt}
  \setlength{\abovecaptionskip}{-2pt}
  \caption{\textbf{Ablations on vital layer selection.}}
  \label{fig:layers_cr}
\end{figure}

Two ablations address this: 1) sharing attention in 10 random non-vital layers (ours-N; 10=$N_v$ vital layers), and 2) sharing with random dropout in all 57 layers, dropping 5/6 to approximate $1-N_v/57$ (ours-D'). Other components remain unchanged.
Results (\cref{fig:layers_cr}) show key detail loss in both settings: ours-N alters hairstyle and removes leg features, while ours-D' shifts clothing color (purple $\to$ red). This confirms vital layers carry critical information. Non-vital layers also influence generation but contain excessive unimportant information—sharing all layers creates a copy-paste effect (fifth column in the \cref{fig:layers_cr}).

% \begin{figure}[h]
%   \centering
%   \vspace{-7pt}
%   \includegraphics[width=\linewidth]{figures/layers_cr.pdf}
%   \caption{\textbf{Ablations on vital layer selection.} 
%   }
%   \label{fig:layers_cr}
%   \vspace{-7pt}
% \end{figure}

\begin{figure}[h]
  \centering
  \vspace{-7pt}
  \includegraphics[width=\linewidth]{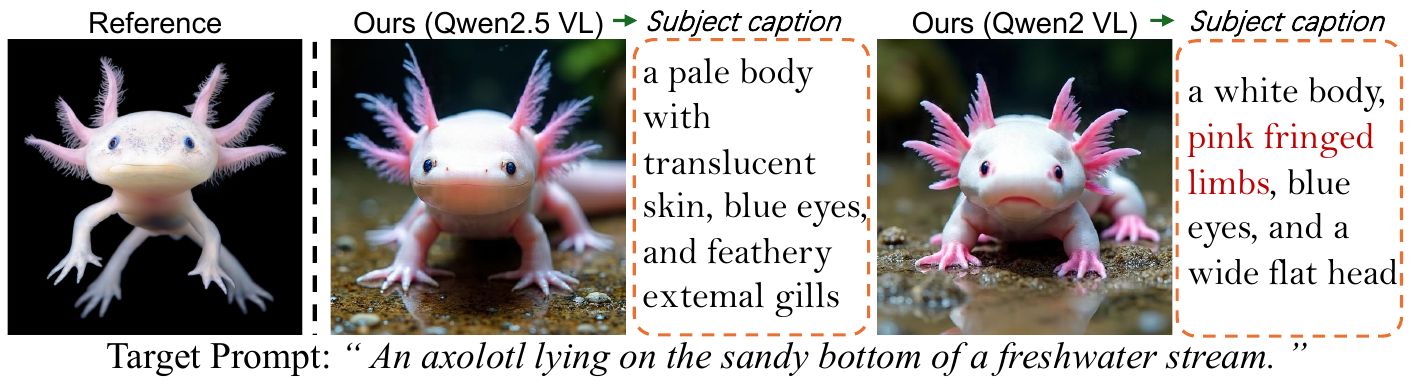}
  \vspace{-7pt}
  \setlength{\abovecaptionskip}{-2pt}
  \caption{\textbf{Stronger MLLMs would yield better results.} }
  \label{fig:rare_subject}
\end{figure}

\myparagraph{Will stronger MLLMs improve our method?} With ongoing advances in MLLMs, our method continues to improve. For example, upgrading from Qwen2-VL to Qwen2.5-VL reduces errors (highlighted in \textcolor{red}{red}) for rare subjects, as illustrated in \cref{fig:rare_subject}.

% \begin{figure}[h]
%   \centering
%   \vspace{-7pt}
%   \includegraphics[width=\linewidth]{figures/rare_subject.pdf}
%   \caption{\textbf{Stronger MLLMs would yield better results.} }
%   \label{fig:rare_subject}
%   \vspace{-7pt}
% \end{figure}

\begin{figure}[h]
    \centering
    \vspace{-7pt}
    \includegraphics[width=\linewidth]{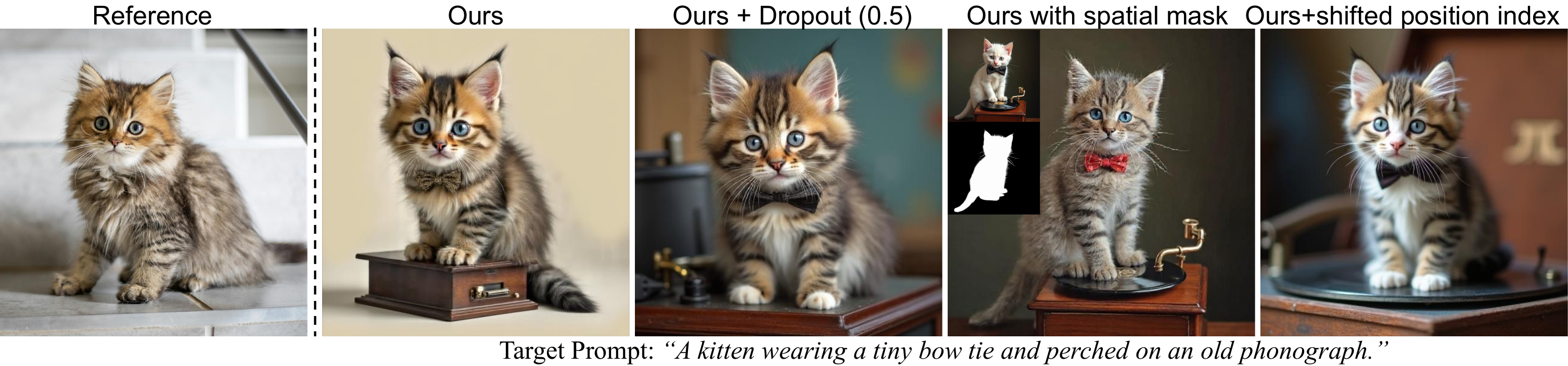}
    \vspace{-7pt}
    \setlength{\abovecaptionskip}{-2pt}
    \caption{\textbf{Strategies to eliminate artifacts.} }
    \label{fig:outline}
    % \lblfig{methoddiagram}
    % \vspace{-7pt}
\end{figure}

\myparagraph{Deeper discussion of artifact mitigation.} We explored two spatial-level strategies: spatial masking (ours-M) and position index shifting of shared attention (ours-S). Both reduce artifacts but introduce trade-offs, as shown in \cref{fig:outline}, ours-S loses details and ours-M misaligns with reference subject's body geometry, lowering quality. We also tried randomly dropping half the shared attention, achieving the best balance and allowing adjustable dropout rates for controlled artifact reduction. Future work will explore adaptive dropout strategies to enhance generalization.

% \begin{figure}[h]
%     \centering
%     \vspace{-7pt}
%     \includegraphics[width=\linewidth]{figures/outline.pdf}
%     \caption{\textbf{Strategies to eliminate artifacts.} }
%     \label{fig:outline}
%     % \lblfig{methoddiagram}
%     \vspace{-7pt}
% \end{figure}

\begin{figure}[h]
    \centering
    \vspace{-7pt}
    \includegraphics[width=\linewidth]{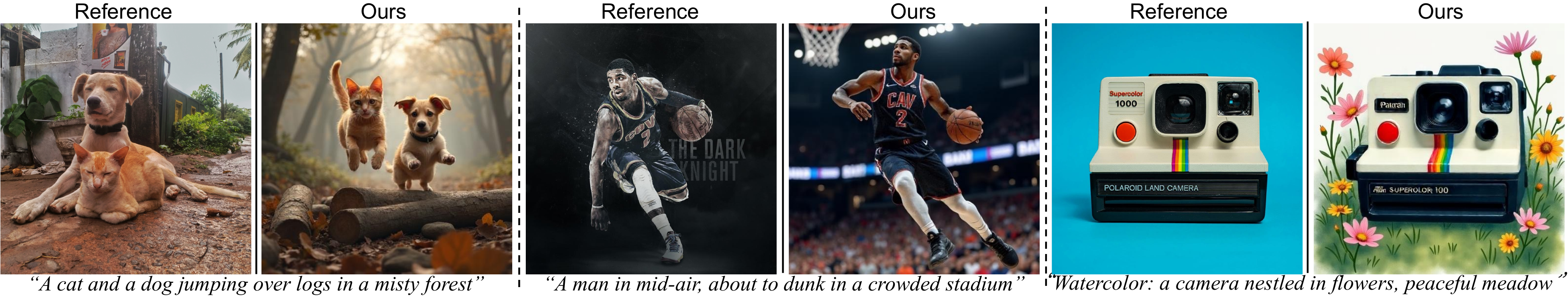}
    \vspace{-7pt}
    \setlength{\abovecaptionskip}{-2pt}
    \caption{\textbf{More qualitative samples.} }
    \label{fig:multi}
    % \vspace{-7pt}
\end{figure}

\myparagraph{More qualitative samples.} \cref{fig:multi} illustrates that our method handles both human subjects (e.g., basketball player) and complex objects (e.g., camera with distinct features), as well as multiple and rare subjects (see in \cref{fig:rare_subject}). While FreeCus is designed for single-subject customization, it can be extended to multi-subject scenes by tailoring the prompts fed into MLLMs.

% \begin{figure}[h]
%     \centering
%     \vspace{-7pt}
%     \includegraphics[width=\linewidth]{figures/multiple_subject2.pdf}
%     \caption{\textbf{More qualitative samples.} }
%     \label{fig:multi}
%     \vspace{-7pt}
% \end{figure}

\myparagraph{Detailed quantitative results on each class.} As shown in \cref{table:all_comparsion}, our genuinely training-free method achieves state-of-the-art or comparable performance across all classes when benchmarked against approaches requiring additional training.

\myparagraph{Prompt for detailed subject caption.} The detailed subject descriptions, discussed in ``Designs for captions'' of \cref{sec:ablation}, are generated by Qwen2-VL with specialized prompts as shown in \cref{fig:detailed_prompt}. 

\begin{figure}[h]
  \centering
  \vspace{-7pt}
  \includegraphics[width=\linewidth]{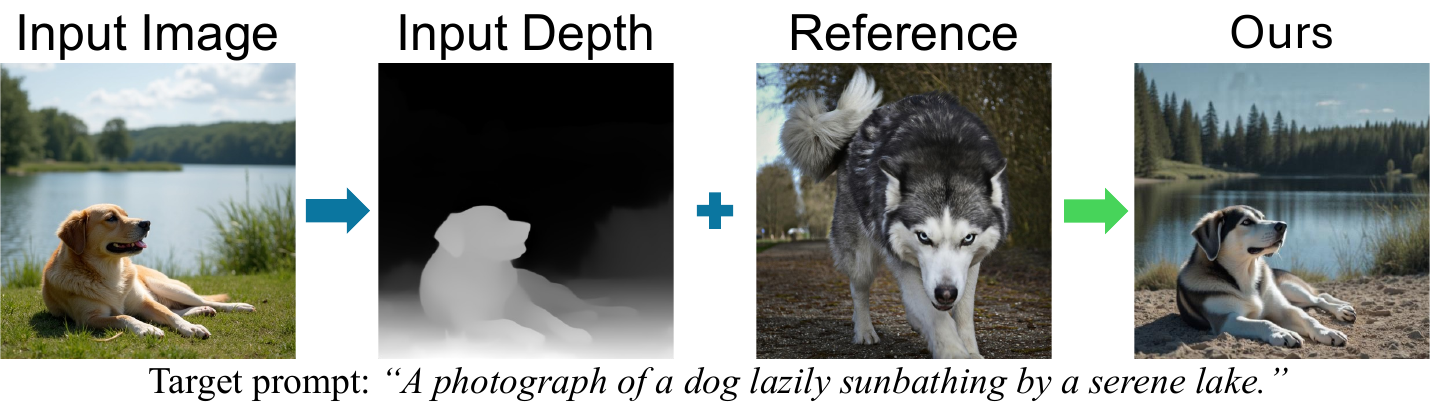}
  \vspace{-7pt}
  \setlength{\abovecaptionskip}{-2pt}
  \caption{\textbf{ Harmonizing with the control model to stabilize target structure.} 
  }
  \label{fig:depth}
\end{figure}
\begin{table}[h]
\centering
\resizebox{\linewidth}{!}{
\begin{tabular}{l|ccc}
Method            & CLIP-T ↑       & CLIP-I ↑       & DINO ↑ \\ \hline
Qwen2VL-Flux      & 0.267          & 0.841          & 0.664  \\
Ours+Qwen2VL-Flux & \textbf{0.274} & \textbf{0.853} & 0.658 
\end{tabular}
}

\caption{\textbf{Quantitative results with and without our method integration in Qwen2VL-Flux framework.}
}
\vspace{-15pt}
\label{table:our_qvlflux}
\end{table}
\begin{table*}[]
\centering
\resizebox{\linewidth}{!}{
\begin{tabular}{lccccccccccccc}
\hline
\multicolumn{1}{c}{}                         &                             & \multicolumn{3}{c}{Animal}                                                                 & \multicolumn{3}{c}{Human}                                                                  & \multicolumn{3}{c}{Object}                                                                 & \multicolumn{3}{c}{Averaged}                                                               \\ \cline{3-14} 
\multicolumn{1}{c}{\multirow{-2}{*}{Method}} & \multirow{-2}{*}{BaseModel} & CLIP-T ↑                     & CLIP-I ↑                     & DINO ↑                       & CLIP-T ↑                     & CLIP-I ↑                     & DINO ↑                       & CLIP-T ↑                     & CLIP-I ↑                     & DINO ↑                       & CLIP-T ↑                     & CLIP-I ↑                     & DINO ↑                       \\ \hline
Textual   Inversion$^\dagger$                & SD v1.5                     & 0.314                        & 0.784                        & 0.537                        & {\color[HTML]{3166FF} 0.281} & 0.645                        & 0.322                        & 0.297                        & 0.709                        & 0.412                        & 0.298                        & 0.713                        & 0.430                        \\
DreamBooth$^\dagger$                         & SD v1.5                     & 0.322                        & 0.817                        & 0.655                        & {\color[HTML]{3166FF} 0.322} & 0.561                        & 0.253                        & {\color[HTML]{3166FF} 0.323} & 0.770                        & 0.568                        & {\color[HTML]{3166FF} 0.322} & 0.716                        & 0.505                        \\
DreamBooth-L$^\dagger$                       & SDXL v1.0                   & {\color[HTML]{3166FF} 0.342} & 0.840                        & 0.724                        & {\color[HTML]{3166FF} 0.339} & 0.623                        & 0.316                        & {\color[HTML]{3166FF} 0.343} & 0.791                        & 0.602                        & {\color[HTML]{3166FF} 0.341} & 0.751                        & 0.547                        \\
BLIP-Diffusion                               & SD v1.5                     & 0.304                        & 0.857                        & 0.692                        & 0.236                        & 0.763                        & 0.567                        & 0.286                        & 0.827                        & 0.658                        & 0.276                        & 0.815                        & 0.639                        \\
Emu2                                         & SDXL v1.0                   & 0.315                        & 0.812                        & 0.621                        & {\color[HTML]{3166FF} 0.284} & 0.736                        & 0.476                        & 0.316                        & 0.742                        & 0.490                        & 0.305                        & 0.763                        & 0.529                        \\
IP-Adapter                                   & SDXL v1.0                   & 0.314                        & 0.892                        & 0.719                        & {\color[HTML]{3166FF} 0.292} & 0.784                        & 0.479                        & 0.307                        & 0.859                        & 0.665                        & 0.305                        & 0.845                        & 0.621                        \\
IP-Adapter-Plus                              & SDXL v1.0                   & 0.293                        & {\color[HTML]{3166FF} 0.939} & {\color[HTML]{3166FF} 0.840} & 0.236                        & {\color[HTML]{3166FF} 0.890} & {\color[HTML]{3166FF} 0.747} & 0.283                        & {\color[HTML]{3166FF} 0.919} & {\color[HTML]{3166FF} 0.834} & 0.271                        & {\color[HTML]{3166FF} 0.916} & {\color[HTML]{3166FF} 0.807} \\
MS-Diffusion                                 & SDXL v1.0                   & {\color[HTML]{3166FF} 0.344} & {\color[HTML]{3166FF} 0.925} & {\color[HTML]{3166FF} 0.816} & {\color[HTML]{3166FF} 0.322} & {\color[HTML]{3166FF} 0.810} & 0.629                        & {\color[HTML]{3166FF} 0.342} & {\color[HTML]{3166FF} 0.885} & {\color[HTML]{3166FF} 0.741} & {\color[HTML]{3166FF} 0.336} & {\color[HTML]{3166FF} 0.873} & {\color[HTML]{3166FF} 0.729} \\
Qwen2VL-Flux                                 & FLUX.1                      & 0.287                        & 0.902                        & 0.704                        & 0.232                        & 0.779                        & 0.669                        & 0.283                        & 0.842                        & 0.619                        & 0.267                        & 0.841                        & 0.664                        \\
IP-Adapter                                   & FLUX.1                      & 0.325                        & 0.898                        & 0.700                        & {\color[HTML]{3166FF} 0.285} & 0.786                        & 0.633                        & {\color[HTML]{3166FF} 0.332} & 0.836                        & 0.581                        & {\color[HTML]{3166FF} 0.314} & 0.840                        & 0.638                        \\ 
OminiControl                                   & FLUX.1                      & {\color[HTML]{3166FF} 0.336}                        & 0.869                        & 0.656                        & {\color[HTML]{3166FF} 0.323} & 0.693                       & 0.439                        & {\color[HTML]{3166FF} 0.331} & 0.829                        & 0.615                        & {\color[HTML]{3166FF} 0.330} & 0.797                        & 0.570                        \\ \hline
Ours                                         & FLUX.1                      & \textbf{0.328}               & \textbf{0.902}               & \textbf{0.738}               & \textbf{0.276}               & \textbf{0.788}               & \textbf{0.675}               & \textbf{0.321}               & \textbf{0.869}               & \textbf{0.677}               & \textbf{0.308}               & \textbf{0.853}               & \textbf{0.696}               \\ \hline
\end{tabular}
}
% \caption{Quantitative evaluation results. 1) Even though the no need of any training resource, our training-free approach outperforms most training-oriented on subject similarity (CLIP-I and DINO indicators), while maintaining good text alignment (CLIP-T indicator). {\color[HTML]{3166FF} Blue} represents the better value by compared methods, and $^\dagger$ represents the optimization-based methods. 2) Ablation study results shows each component has an important impact on the subject similarity. {\color[HTML]{CE6301} Brown} represents the better value by ablation methods.}
% \caption{\textbf{Quantitative evaluation results.} (1) Despite without any training resources, our zero-shot approach outperforms most training-oriented methods in subject similarity (CLIP-I and DINO scores), while maintaining strong text alignment (CLIP-T score). {\color[HTML]{3166FF} Blue} represents higher scores than our approach by compared methods, and \(^{\dagger}\) marks optimization-based methods. (2) Ablation study results demonstrate that each component significantly impacts subject similarity.}
\caption{\textbf{Quantitative evaluation results for each class.} {\color[HTML]{3166FF} Blue} indicates scores higher than ours, and \(^{\dagger}\) denotes optimization-based methods. 
%(1) Our approach, requiring no training resources, surpasses most training-based methods in subject similarity (CLIP-I and DINO) while preserving text alignment (CLIP-T). (2) Ablation studies confirm each component's significant contribution to subject similarity.
}
\vspace{-15pt}
\label{table:all_comparsion}
\end{table*}

\myparagraph{Prompt for style transfer.} For the style transfer task, the prompt fed to Qwen2-VL is ``Describe this style briefly and precisely in max 20 words, focusing on its aesthetic qualities, visual elements, and distinctive artistic characteristics.''.

Subsequently, the prompt fed to Qwen2.5 is ``Please extract only the stylistic and artistic characteristics of the style from this description, removing any information about physical objects, specific subjects, narrative elements, or factual content. Focus solely on the aesthetic qualities, visual techniques, artistic movements, and distinctive style elements. Return only the extracted style description without any additional commentary. The description is: \{ [output from Qwen2-VL] \}''.

\myparagraph{Quantitative results with and without our method integration in DiT-based framework.} As shown in \cref{table:our_qvlflux}, compared to the original Qwen2VL-Flux, our method combined with it achieves higher scores on two metrics, further demonstrating the compatibility and orthogonality of \textit{FreeCus} with other DiT-based models.

\myparagraph{Subject-driven layout-guidance generation.} As illustrated in \cref{fig:depth}, our method also supports layout-guided synthesis when integrated with the Flux.1-Depth-dev model.

% \begin{figure}[h]
%   \centering
%   \includegraphics[width=\linewidth]{figures/depth.pdf}
%   \caption{\textbf{ Harmonizing with the control model to stabilize target structure.} 
%   }
%   \label{fig:depth}
%   \vspace{-15pt}
% \end{figure}

\section{Compared Methods and Implementation Details}
% \subsection{Details of Compared Methods}

\myparagraph{IP-Adapter (IPA) \cite{ye2023ipa}} IPA introduces a lightweight adapter that decouples image and text features, addressing limitations in fine-grained control when merging these features in cross-attention layers. For IPA (Flux.1) implementation, we use the third-party code from \href{https://github.com/XLabs-AI/x-flux}{XLabs-AI}.

\myparagraph{MS-Diffusion (MS-D) \cite{msDiffusion}} MS-D incorporates grounding tokens with feature resampling to preserve subject detail fidelity. It requires inputting a bounding box for layout guidance; we set the default box values to [0.25, 0.25, 0.75, 0.75].

\myparagraph{Qwen2VL-Flux (QVL-Flux) \cite{erwold-2024-qwen2vl-flux}} QVL-Flux replaces Flux's conventional T5-XXL text encoder with a vision-language model, enabling image-to-image generation. We utilize the official repository and weights to generate $1024 \times 1024$ images.

\myparagraph{Textual Inveresion (TI) \cite{textualInversion}} TI updates only the new token embedding representing the novel subject while keeping all other parameters frozen. Experimental results are from the DreamBench++ \cite{peng2024dreambench++} implementation.

\myparagraph{DreamBooth \cite{ruiz2023dreambooth}} DreamBooth updates all layers of the T2I model to maintain visual fidelity and employs prior preservation loss to prevent language drift. DreamBooth-Lora only updates additional lora adapters. Experimental results are from the DreamBench++ \cite{peng2024dreambench++} implementation.

\myparagraph{BLIP-Diffusion (BLIP-D) \cite{li2024blipDiffusion}} BLIP-D leverages the pretrained BLIP-2 multimodal encoder to create multiple learnable embeddings representing input subject features, then fine-tunes the base model to adapt these embeddings for personalization. Experimental results are from the DreamBench++ \cite{peng2024dreambench++} implementation.

\myparagraph{Emu2 \cite{sun2024generativeEmu2}}  Emu2 employs an autoregressive approach to process multimodal information with a predict-the-next-element objective. Images are tokenized via a visual encoder and interleaved with text tokens, enabling straightforward customization with target text. Experimental results are from the DreamBench++ \cite{peng2024dreambench++} implementation.

\myparagraph{OminiControl \cite{tan2024ominicontrol}} OminiControl performs multiple image-to-image tasks using a unified sequence processing strategy and dynamic position encoding, introducing only lightweight trainable LoRA parameters. We reproduced results using the official repository.

% \clearpage

\begin{prompt}
[Task Description]
As an experienced image analyst, your task is to provide a detailed description of the main features and characteristics of the given {} in this image according to the following criteria.

[Feature Analysis Criteria]
Analyze and describe the following visual elements:
1. Shape
- Main body outline
- Overall structure
- Proportions and composition
- Spatial organization

2. Color
- Color palette and schemes
- Saturation levels
- Brightness/contrast
- Color distribution patterns

3. Texture
- Surface qualities
- Detail clarity
- Visual patterns
- Material appearance

4. Subject-Specific Features
- If human/animal: facial features, expressions, poses
- If object: distinctive characteristics, condition
- If landscape: environmental elements, atmosphere

[Description Quality Levels]
Your description should aim for the highest level of detail:
Level 1: Basic identification of main elements
Level 2: Description of obvious features
Level 3: Detailed analysis of multiple characteristics
Level 4: Comprehensive analysis with subtle details

[Output Format]
Please provide your analysis in the following structure:

Main Subject: [Brief identifier]
Primary Features:
- Shape: [Description]
- Color: [Description]
- Texture: [Description]
- Subject-Specific Details: [Description]
Overall Composition: [Brief summary]
\end{prompt}
\captionof{figure}{Prompt for Detailed Subject Caption.}
\label{fig:detailed_prompt}

\end{document}